# Tiered Latent Representations and Latent Spaces for Molecular Graphs

Daniel T. Chang (张遵)

*IBM (Retired)* dtchang43@gmail.com

**Abstract:**

Molecular graphs generally contain subgraphs (known as groups) that are identifiable and significant in composition, functionality, geometry, etc. Flat latent representations (node embeddings or graph embeddings) fail to represent, and support the use of, groups. Fully hierarchical latent representations, on the other hand, are difficult to learn and, even if learned, may be too complex to use or interpret. We propose tiered latent representations and latent spaces for molecular graphs as a simple way to explicitly represent and utilize groups, which consist of the atom (node) tier, the group tier and the molecule (graph) tier. Specifically, we propose an architecture for learning tiered latent representations and latent spaces using graph autoencoders, graph neural networks, differentiable group pooling and the membership matrix. We discuss its various components, major challenges and related work, for both a deterministic and a probabilistic model. We also briefly discuss the usage and exploration of tiered latent spaces. The tiered approach is applicable to other types of structured graphs similar in nature to molecular graphs.

## 1 Introduction

*Molecular graphs* [1] are widely used for representing molecular structure. They are of increasing importance due to their flexibility and power and the rapid advances in *graph neural networks (GNNs)* [6-7]. Molecular graphs generally contain subgraphs (known as *groups*) that are identifiable and significant in composition, functionality, geometry, etc. Flat latent representations (node embeddings or graph embeddings) fail to represent, and support the use of, groups. Fully hierarchical latent representations [2], on the other hand, are difficult to learn and, even if learned, may be too complex to use or interpret.

We propose *tiered latent representations and latent spaces* for molecular graphs as a simple way to explicitly represent and utilize groups, which consist of the *atom (node) tier*, the *group tier* and the *molecule (graph) tier*. Specifically, we propose an architecture for learning tiered latent representations and latent spaces using *graph autoencoders (GAEs)* [7, 11-12], GNNs, *differentiable group pooling (DiffGroupPool)* and the membership matrix.

We discuss the various components, major challenges and related work of the tiered approach, for both a deterministic and a probabilistic model. We also briefly discuss the usage and exploration of tiered latent spaces. The tiered approach is applicable to other types of structured graphs similar in nature to molecular graphs.

# 2 Preliminaries

## 2.1 Notations

For variables, we use bold uppercase characters to denote *matrices* and bold lowercase characters to denote *vectors*, e.g. **W** and **w**. The transpose of matrix is denoted as $\mathbf{W}^T$. A *graph* is represented as $G = (V, E)$ where $V = \{v_1, \ldots, v_N\}$ is a set of $N = |V|$ *nodes* (*vertices*) and $E \subseteq V \times V$ is a set of $U = |E|$ *edges* between nodes. We use $\mathbf{A} \in \mathbb{R}^{N \times N}$ to denote the *adjacency matrix*, where its ith row, jth column and an element denoted as $\mathbf{A}(i, :)$, $\mathbf{A}(:, j)$ and $\mathbf{A}(i, j)$, respectively. We use $\mathbf{F}^V \in \mathbb{R}^{N \times d}$ and $\mathbf{F}^E \in \mathbb{R}^{U \times s}$ to denote *features* for nodes and edges respectively. *Functions* are marked by Algerian font, e.g. $\mathbb{F}(.)$.

We use **Z** to denote *embeddings (latent representations)*, **X** to denote *input node embeddings*, and **H** to denote hidden, intermediate node embeddings. We use **M** to denote the *membership matrix*. For the probabilistic model, we use $\boldsymbol{\mu}_Z$ and $\boldsymbol{\sigma}_Z$ to denote the *sufficient statistics* of latent variable distributions [2], which are the mean and standard deviation for Gaussian distributions.

When needed, {} is used to represent subscript. So, N{t} is the same as $N_t$.

## 2.2 Molecular Graphs

The *molecular graph* [1] represents a *valence bond model* of molecule structure, which has proven to be an incredibly useful model for chemistry. A valence bond model is a way of allocating a molecule's nuclei and electrons into *atoms* and *bonds* in a way that makes sense. The function of a molecular graph is to clearly represent a particular valence bond model, not dictate which one should be used.

In a molecular graph, *nodes* correspond to atoms and *edges* correspond to bonds. Molecular graphs commonly have node features and edge features. The *node (atom) features* [8-9] typically consists of:

- atom type, hybridization, hydrogen bonding, aromaticity, ring sizes, formal charge, partial charge, chirality,

and the *edge (bond) features* [8] typically include:

- bond type (single, double, triple or aromatic), conjugated bond, same ring.

*Groups*

Groups are a key aspect of molecular structure and, therefore, molecular graphs. A *group* is a collection of atoms at a site within a molecule with a *common bonding pattern*. The group reacts in a typical way, generally independent of the rest of the molecule. Groups give the molecule its *properties*, regardless of what molecule contains it; they are centers of *chemical reactivities*. Exemplary types of groups include *functional groups* [1] and *R-groups* (radical groups), though a group can simply be two or more atoms bound together as a single unit.

As an example, the following is the molecular structure (graph) of *vanillin*, which is the primary taste behind vanilla, that has groups:

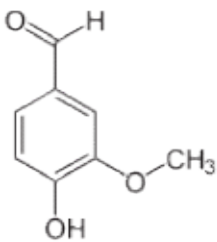

The groups are: a carbonyl group (-CH=O), a methoxy group (-O-$CH_3$) and a hydroxyl group (-OH), which are attached to the core benzene group (---$C_6H_3$).

## 2.3 Node Embedding

The goal of *node embedding* [5] is to encode nodes as low-dimensional continuous latent representations, $z^V \in R^d$, that summarize their graph position and the structure of their local graph neighborhood. The node embeddings can be viewed as encoding nodes into a *latent space*, where geometric relations in this latent space correspond to edges and other correlations in the original graph.

## 2.4 Subgraph Embedding

The goal of *(sub)graph embedding* [5] is to learn a continuous latent representation, $z^S \in R^d$, of an *induced subgraph G[S]* of the full graph G, where S ⊆ V. (The induced subgraph G[S] is a graph whose vertex set is S and whose edge set consists of all of the edges in E that have both endpoints in S.) The embedding, $z^S$, can then be used to make predictions about the entire subgraph and, in the case of a tiered architecture, as discussed in the next section, to generate the *graph embedding $z^G$*.

Subgraph embedding can be viewed as direct extensions of node embedding. A subgraph consists of a set of connected nodes. Therefore, subgraph embeddings can be generated by *aggregating* sets of node embeddings correspond to subgraphs [5]. The simplest way to generate subgraph embeddings is by *summing* all the individual node embeddings in the subgraph, as in *neural molecular descriptors* [1, 5, 8], which is called *sum pooling* in the context of graph pooling [7]:

$$\mathbf{z}^S = \sum_{V_i \in S} \mathbf{z}^V_i$$

### 2.5 Group Embedding

*Groups* are induced subgraphs that are identifiable and significant in composition, functionality, geometry, etc. To generate *group embeddings*, we partition a graph into its *optimal set* of groups. Take the example of *vanillin*, its optimal set of groups consists of: a carbonyl group (-CH=O), a methoxy group (-O-CH$_3$), a hydroxyl group (-OH), and the core benzene group (---C$_6$H$_3$).

Whereas node embedding and graph embedding are well defined and well studied, group embedding is *novel*. There are major challenges, particularly how to *define and identify* groups.

## 3 Tiered Latent Representations and Latent Spaces

We propose *tiered latent representations and latent spaces* for molecular graphs as a simple way to explicitly represent and utilize groups, which consist of, from bottom to top: the *atom (node)* tier, the *group* tier, and the *molecule (graph)* tier. Specifically, we propose the following *tiered GAE architecture* for learning tiered latent representations and latent spaces:

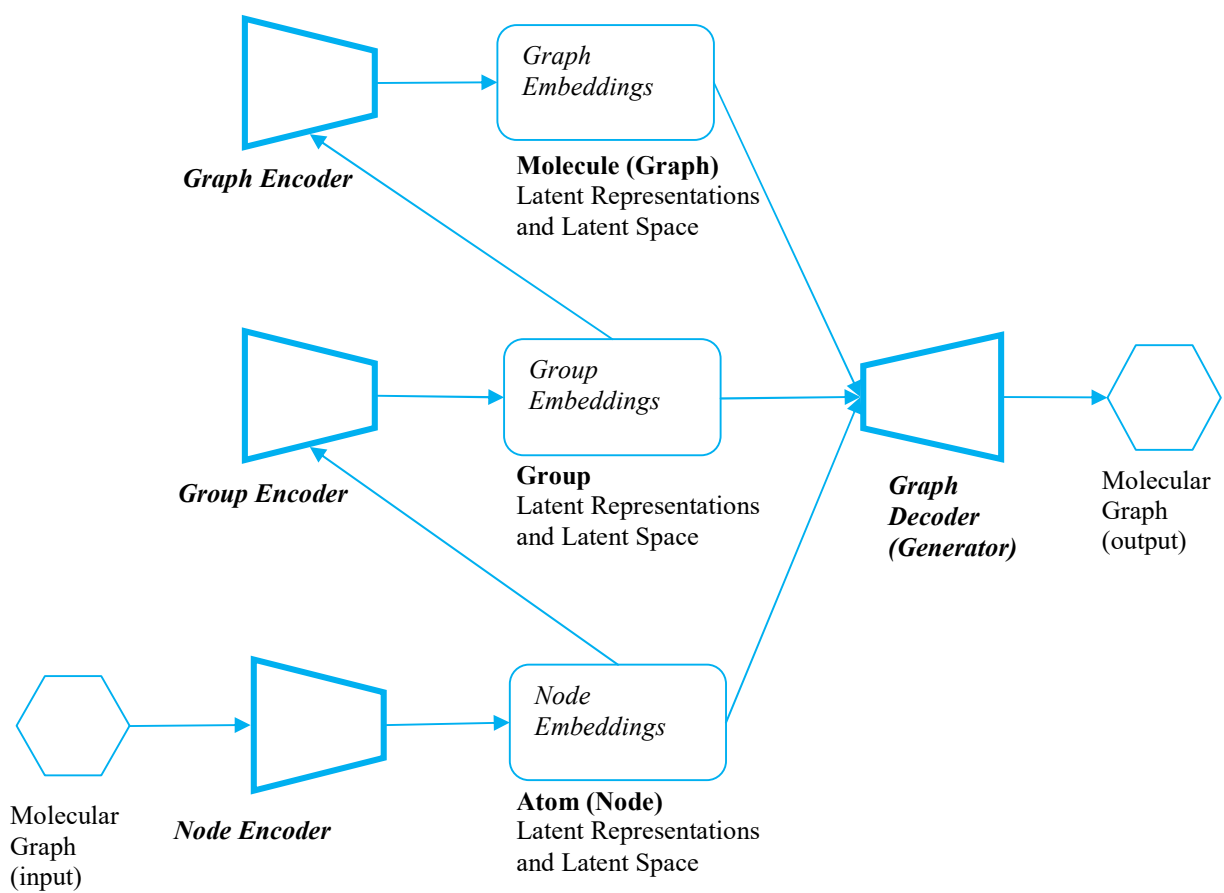

It is important to note that the three tiers of embeddings are jointly learned / trained. They thus form a three-tier associated hierarchical embeddings for each input graph, from the bottom tier to the top tier: node embeddings -> group embedding(s) -> graph embedding. This is achieved by using GNNs, *DiffGroupPool* and the membership matrix, as discussed in the next section.

The tiered GAE architecture is an excellent example of *concept-oriented deep learning* [4] and *generative concept representations* [3]. Each embedding tier embeds the generative concept representations of an important *chemical concept* and its subconcepts. The node embedding tier embeds the chemical concept *atom* and its subconcepts (*subclasses of atom* or *chemical elements*) such as carbon, hydrogen, oxygen, etc. The group embedding tier embeds the chemical concept *group* and its subconcepts (*subclasses of group*) such as carbonyl group, methoxy group, hydroxyl group and benzene group. The graph embedding tier embeds the chemical concept *molecule* and its subconcepts (*subclasses of molecule*) such as vanillin.

The tiered approach is general and applicable to other types of *structured graphs* similar in nature to molecular graphs. It is also flexible, allowing the user to *control tiering for usage and interpretability* by having one or more middle tiers (group, super group …) and what each embeds.

The tiered GAE architecture can be realized in a deterministic or a probabilistic model. We first discuss the deterministic model, the *tiered GAE model*. The probabilistic model, the *tiered variational GAE (VGAE) model*, which extends the tiered GAE model, is discussed in 6 Tiered VGAE Model.

## 4 Tiered GAE Model

In the *tiered GAE model*, we use GNNs, *DiffGroupPool* and the membership matrix to learn and generate, jointly, node embeddings, group embeddings and graph embeddings. *DiffGroupPool* is based on *DiffPool* [10] with the following major extensions and modifications:

- DiffPool is designed for L-layer hierarchical clustering and embedding, where L can be any number ≥ 3. *DiffGroupPool* is tailored for T-tier *grouping and embedding*, where T is a small number ≥ 3. We use T = 3.
- DiffPool uses a clustering GNN at each layer to learn the assignment matrix, based on the adjacency matrix and node features. *DiffGroupPool* uses a deterministic algorithm to learn the *membership matrix*, based on functional groups and simple aromatic rings.

### 4.1 GNNs

Following DiffPool, we consider GNNs that employ the following "message-passing" architecture, using node embedding as an example:

$$\mathbf{H}^{(k)} = \mathbf{MP}(\mathbf{A}, \mathbf{H}^{(k-1)}; \theta^{(k)})$$

where $\mathbf{H}^{(k)} \in \mathbf{R}^{N \times d}$ are the *hidden, intermediate node embeddings* (i.e., "messages") computed after *k iterations* of the GNN and $\mathbf{MP}$ is the *message propagation function*, which depends on the adjacency matrix $\mathbf{A}$, trainable parameters $\theta^{(k)}$, and the node embeddings $\mathbf{H}^{(k-1)}$. The input node embeddings $\mathbf{H}^{(0)}$ are initialized using the node features on the graph, $\mathbf{H}^{(0)} = \mathbf{F}^V$. A full GNN will run K iterations to generate the output *node embeddings* $\mathbf{Z}^V = \mathbf{H}^{(K)} \in \mathbf{R}^{N \times d}$, where K is usually in the range 2-6.

For generalization and simplicity, we use

$$\mathbf{Z} = \mathbf{GNN}(\mathbf{A}, \mathbf{X})$$

to denote the embeddings generated using a GNN that implements K iterations of message passing according to input adjacency matrix **A** and input 'node' embeddings **X**. The embeddings could be node embeddings, group embeddings or graph embeddings.

For each tier t we thus have ($\mathbf{Z}^{(1)}$: node embeddings, $\mathbf{Z}^{(2)}$: group embeddings, and $\mathbf{Z}^{(3)}$: graph embeddings):

$$\mathbf{Z}^{(t)} = \text{GNN}(\mathbf{A}^{(t)}, \mathbf{X}^{(t)})$$

with $\mathbf{A}^{(1)} = \mathbf{A}$ and $\mathbf{X}^{(1)} = \mathbf{F}^V$.

## 4.2 Stacking GNNs and DiffGraphPool Modules

Given $\mathbf{Z} = \text{GNN}(\mathbf{A}, \mathbf{X})$, the output of a GNN, and an adjacency matrix $\mathbf{A} \in \mathbb{R}^{N \times N}$, we seek to output a new *coarsened graph* containing M < N nodes, with weighted adjacency matrix $\mathbf{A}' \in \mathbb{R}^{M \times M}$ and node embeddings $\mathbf{Z}' \in \mathbb{R}^{M \times d}$. This new coarsened graph can then be used as input to another GNN tier, and this whole process can be repeated T times, generating a model with *T GNN tiers*. To do this, at each tier, we first run a GNN to obtain node embeddings. We then use these node embeddings to *group* nodes together as coarsened nodes using *DiffGroupPool*. This is illustrated in the following (the *DiffGroupPool* module, DGP(.), is discussed in the next subsection):

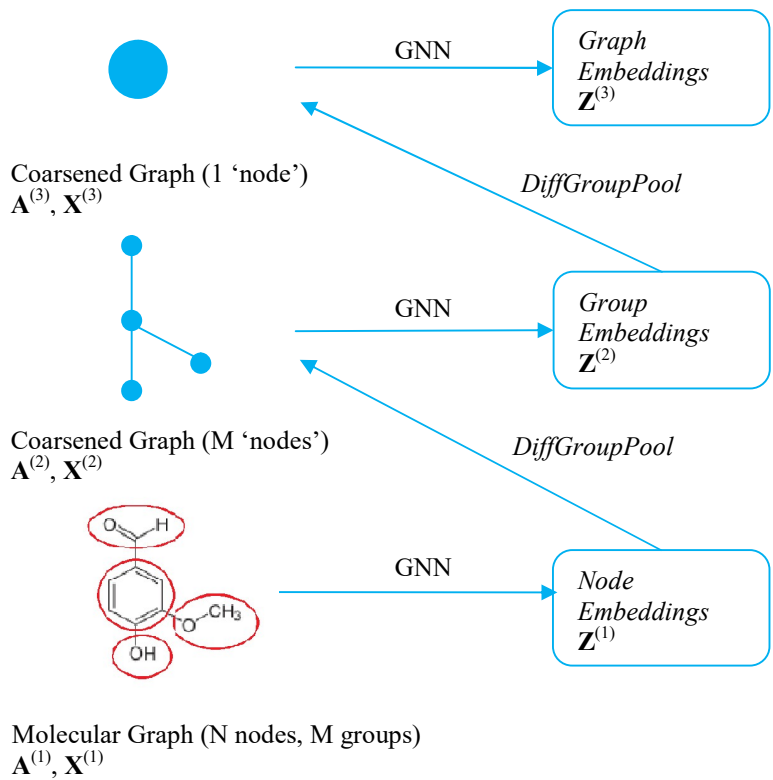

From the perspective of the tiered GAE model, we have the following *tiered encoders* with stacked GNNs and *DiffGroupPool* modules, from bottom to top:

- Node encoder: $\text{GNN}(\mathbf{A}^{(1)}, \mathbf{X}^{(1)}) \Rightarrow \mathbf{Z}^{(1)}$

- Group encoder: $\text{DGP}(\mathbf{A}^{(1)}, \mathbf{Z}^{(1)}) \Rightarrow (\mathbf{A}^{(2)}, \mathbf{X}^{(2)}), \text{GNN}(\mathbf{A}^{(2)}, \mathbf{X}^{(2)}) \Rightarrow \mathbf{Z}^{(2)}$

- Graph encoder: $\text{DGP}(\mathbf{A}^{(2)}, \mathbf{Z}^{(2)}) \Rightarrow (\mathbf{A}^{(3)}, \mathbf{X}^{(3)}), \text{GNN}(\mathbf{A}^{(3)}, \mathbf{X}^{(3)}) \Rightarrow \mathbf{Z}^{(3)}$

## 4.3 Differentiable Pooling via the Membership Matrix

We denote the *membership matrix* at tier t as $\mathbf{M}^{(t)} \in \mathbb{R}^{N\{t\} \times N\{t+1\}}$. Each row of $\mathbf{M}^{(t)}$ corresponds to one of the N{t} *nodes* at tier t, and each column of $\mathbf{M}^{(t)}$ corresponds to one of the N{t+1} *groups* at the next coarsened tier t+1. $\mathbf{M}^{(t)}$ provides membership assignments of each node at tier t to a group in the next coarsened tier t+1. The membership matrix is discussed in the next section. For now we assume it already exists.

We denote the input adjacency matrix as $\mathbf{A}^{(t)}$ at tier t and the input node embeddings as $\mathbf{Z}^{(t)}$. Given these, the *DiffGroupPool* module for tier t+1 is denoted as:

$$(\mathbf{A}^{(t+1)}, \mathbf{X}^{(t+1)}) = \text{DGP}(\mathbf{A}^{(t)}, \mathbf{Z}^{(t)})$$

It coarsens the input graph, generating a new coarsened adjacency matrix $\mathbf{A}^{(t+1)}$ and a new matrix of coarsened embeddings $\mathbf{X}^{(t+1)}$ in this coarsened graph. In particular, *DiffGroupPool* applies the following equations:

$$\mathbf{X}^{(t+1)} = (\mathbf{M}^{(t)})^T \mathbf{Z}^{(t)} \in \mathbb{R}^{N\{t+1\} \times d}$$

$$\mathbf{A}^{(t+1)} = (\mathbf{M}^{(t)})^T \mathbf{A}^{(t)} \mathbf{M}^{(t)} \in \mathbb{R}^{N\{t+1\} \times N\{t+1\}}$$

The first equation takes the node embeddings $\mathbf{Z}^{(t)}$ and aggregates these embeddings according to the membership matrix $\mathbf{M}^{(t)}$, generating embeddings for each of the $N\{t+1\}$ groups. Similarly, the second equation takes the adjacency matrix $\mathbf{A}^{(t)}$ and generates a coarsened adjacency matrix denoting the connectivity strength between each pair of groups.

Note that $\mathbf{A}^{(t+1)}$ represents a fully connected edge-weighted graph; each entry $\mathbf{A}^{(t+1)}(i, j)$ can be viewed as the connectivity strength between group i and group j. Similarly, the i-th row of $\mathbf{X}^{(t+1)}$ corresponds to the input embeddings of group i. Together, the coarsened adjacency matrix $\mathbf{A}^{(t+1)}$ and group embeddings $\mathbf{X}^{(t+1)}$ can be used as input to another, t+1, GNN tier.

### 4.4 Graph Decoder (Generator)

We denote the *tiered graph decoder (generator)* as:

$$(\hat{\mathbf{A}}, \hat{\mathbf{X}}) = \text{DEC}(\mathbf{Z}^{(3)}, \mathbf{Z}^{(2)}, \mathbf{Z}^{(1)}),$$

where $\hat{\mathbf{A}}$ is the reconstructed adjacency matrix and $\hat{\mathbf{X}}$ is the reconstructed node features. The tiered GAE model is trained by minimizing the *reconstruction loss function*:

$$\text{LOSS}((\hat{\mathbf{A}}, \hat{\mathbf{X}}), (\mathbf{A}, \mathbf{X})),$$

and the parameters in the *tiered GNNs* are optimized by *stochastic gradient descent*.

Currently, most graph decoders are pairwise decoders, which decode pairwise relations between nodes. Developing decoding algorithms that are capable of decoding complex (sub)graphs is an important direction for future work [5]. We discuss some recent work in the following.

*GAE*

The *GAE* [11] generates node embeddings **Z** using a *graph convolutional network (GCN)*:

$$\mathbf{Z} = \text{GCN}(\mathbf{A}, \mathbf{X}).$$

Its pairwise decoder reconstructs the adjacency matrix $\hat{\mathbf{A}}$ as follows:

$$\hat{\mathbf{A}} = \sigma(\mathbf{Z}\mathbf{Z}^T),$$

where $\sigma(.)$ is the logistic sigmoid function.

*MC-GC*

The *MC-GC* [12] also generates node embeddings **Z** using a GCN:

$$\mathbf{Z} = \text{GCN}(\mathbf{A}, \mathbf{X}).$$

Its decoder reconstructs the adjacency matrix $\hat{\mathbf{A}}$ as:

$$\hat{\mathbf{A}} = \mathbf{Z}\boldsymbol{\Theta}_{dec}\mathbf{Z}^T,$$

where $\boldsymbol{\Theta}_{dec}$ are parameters for the decoder.

## 5 Membership Matrix

For a T-tiered architecture there are T-1 membership matrices. The membership matrix at tier T-1, $\mathbf{M}^{(T-1)}$, is known since there is only one group at tier T, the graph. Therefore, all "nodes" at tier T-1 are members of the graph.

With T = 3, there are 2 membership matrices: $\mathbf{M}^{(1)}$ the *group membership matrix*, and $\mathbf{M}^{(2)}$ the *graph membership matrix*. Since $\mathbf{M}^{(2)}$ is known, we only need to generate $\mathbf{M}^{(1)}$, with each row of $\mathbf{M}^{(1)}$ corresponding to one of the *N nodes* and each column of $\mathbf{M}^{(1)}$ corresponding to one of the *M groups*. To do so, we need to partition a graph into its *optimal set* of groups and generate $\mathbf{M}^{(1)}$ based on that.

### 5.1 Group Membership

*Groups* are induced subgraphs that are identifiable and significant in composition, functionality, geometry, etc. However, their composition, functionality, geometry, etc. vary greatly and there is no precise definition of groups. All current software systems [13] to identify groups in molecular structures (graphs) are based on a *predefined list* of molecular substructures (subgraphs).

Using *edge (bond) features* [8], we can infer the following information about *group membership* for a molecular graph:

- bond type (single, double, triple or aromatic): nodes bonded by double, triple or aromatic bonds should be members of the same group,
- conjugated bond: nodes bonded by conjugated bonds should be members of the same group,
- same ring: nodes bonded in the same ring should be members of the same group.

The above information, however, is not sufficient to identify the *optimal set* of groups for a molecular graph.

### 5.2 Automatic Identification of Functional Groups

An algorithm to automatically identify all *functional groups (FGs)* in a molecular graph based on iterative marching through its nodes (atoms) is described in [14]. Since the majority of FGs contain *heteroatoms*, which are atoms in the ring of a cyclic compound other than carbon atoms, the algorithm is based on processing heteroatoms and their environment with the addition of some other functionalities, like multiple carbon–carbon bonds. When extracting FGs from the bioactive portion of the ChEMBL database, the algorithm results in identification of 3080 unique FGs.

### 5.3 Graph Partitioning

Given that all FGs in a molecular graph can be automatically identified, as described above, we propose the following algorithm to partition a graph into its *optimal set* of groups based on FGs and simple aromatic rings:

- Identify all *FGs* in the graph as $FG_1, \ldots, FG_I$.
- Identify all *simple aromatic rings* (each including attached hydrogen atoms) in the graph as $AG_1, \ldots, AG_J$.
- Extract all FGs and simple aromatic rings, and remove their nodes from the graph.
- The rest of the graph should contain K *connected components*: $CG_1, \ldots, CG_K$.

- The *optimal set* of groups consist of: $FG_1, \ldots, FG_I, AG_1, \ldots, AG_J$ and $CG_1, \ldots, CG_K$. The number of groups in the optimal set is $M = I + J + K$.

With the resulting optimal set of groups, it is straightforward to generate $\mathbf{M}^{(1)}$. If a node is a member of m groups (e.g., a FG and a AG), its *membership strength* is $\frac{1}{m}$ in each group. In general, most, if not all, nodes should be a member of a single group and have a membership strength of 1 in that group.

As an example, using the algorithm, *vanillin* is partitioned into its optimal set of groups: a carbonyl group (-CH=O), a methoxy group ($-O-CH_3$), a hydroxyl group (-OH), and the core benzene group ($---C_6H_3$):

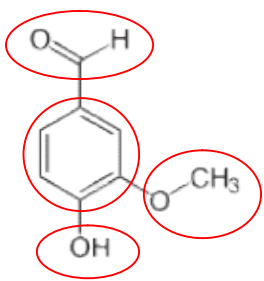

# 6 Tiered VGAE Model

In the *tiered VGAE model*, we extend the deterministic *tiered GAE model* into a probabilistic model by extending its components: GNNs, *DiffGroupPool* and graph decoder, to work together as *variational autoencoders (VAEs)* [3].

We denote the *encoder (inference) model* as [2]:

$$q_\varphi(\mathbf{Z} \mid \mathbf{A}, \mathbf{X}) = q_\varphi(\mathbf{Z}^{(1)} \mid \mathbf{A}, \mathbf{X})\, q_\varphi(\mathbf{Z}^{(2)} \mid \mathbf{Z}^{(1)})\, q_\varphi(\mathbf{Z}^{(3)} \mid \mathbf{Z}^{(2)})$$

## 6.1 GNNs

The parameters for the encoder (inference) model are specified using a GNN:

$$(\boldsymbol{\mu}_Z, \boldsymbol{\sigma}_Z) = \mathrm{GNN}_\varphi(\mathbf{A}, \mathbf{X})$$

where $\boldsymbol{\mu}_Z$ and $\boldsymbol{\sigma}_Z$ denote the *sufficient statistics* for the variational marginals.

For each tier t we have:

$$(\boldsymbol{\mu}_Z^{(t)}, \boldsymbol{\sigma}_Z^{(t)}) = \text{GNN}_\varphi(\mathbf{A}^{(t)}, \mathbf{X}^{(t)}),$$

with $\mathbf{A}^{(1)} = \mathbf{A}$ and $\mathbf{X}^{(1)} = \mathbf{F}^V$.

## 6.2 Stacking GNNs and DiffGraphPool Modules

Same as the tiered GAE model, given the output of a GNN (in this case we use $\boldsymbol{\mu}_Z$ in place of $\mathbf{Z}$) and an adjacency matrix $\mathbf{A}$, we seek to output a new *coarsened graph* containing M < N nodes, with weighted adjacency matrix $\mathbf{A}'$ and node embeddings $\mathbf{Z}'$. This new coarsened graph can then be used as input to another GNN tier, and this whole process can be repeated T times, generating a model with *T GNN tiers*. This is illustrated in the following (the *DiffGroupPool* module, $\text{DGP}(.)$, is discussed in the next subsection):

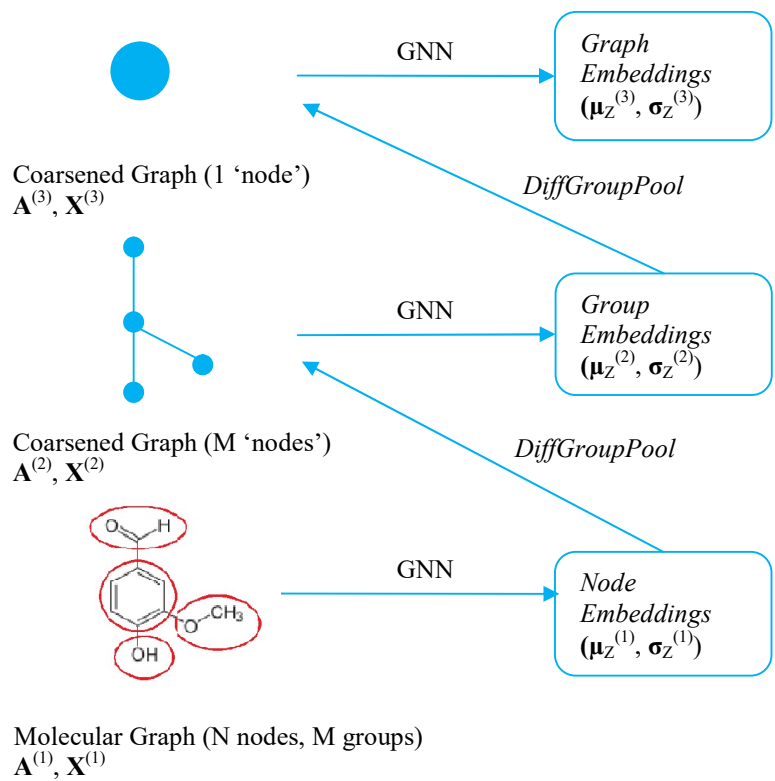

From the perspective of the tiered VGAE model, we have the following *tiered variational encoders* with stacked GNNs and *DiffGroupPool* modules, from bottom to top:

- Node encoder: $\text{GNN}_\varphi(\mathbf{A}^{(1)}, \mathbf{X}^{(1)}) \Rightarrow (\boldsymbol{\mu}_Z^{(1)}, \boldsymbol{\sigma}_Z^{(1)})$

- Group encoder: $\text{DGP}(\mathbf{A}^{(1)}, \boldsymbol{\mu}_Z^{(1)}) \Rightarrow (\mathbf{A}^{(2)}, \mathbf{X}^{(2)})$, $\text{GNN}_\varphi(\mathbf{A}^{(2)}, \mathbf{X}^{(2)}) \Rightarrow (\boldsymbol{\mu}_Z^{(2)}, \boldsymbol{\sigma}_Z^{(2)})$

- Graph encoder: $\text{DGP}(\mathbf{A}^{(2)}, \boldsymbol{\mu}_Z^{(2)}) \Rightarrow (\mathbf{A}^{(3)}, \mathbf{X}^{(3)})$, $\text{GNN}_\varphi(\mathbf{A}^{(3)}, \mathbf{X}^{(3)}) \Rightarrow (\boldsymbol{\mu}_Z^{(3)}, \boldsymbol{\sigma}_Z^{(3)})$

## 6.3 Differentiable Pooling via the Membership Matrix

The *DiffGroupPool* module for tier t+1 is denoted as (replacing $\mathbf{Z}^{(t)}$ with $\boldsymbol{\mu}_Z^{(t)}$):

$$(\mathbf{A}^{(t+1)}, \mathbf{X}^{(t+1)}) = \text{DGP}(\mathbf{A}^{(t)}, \boldsymbol{\mu}_Z^{(t)})$$

It coarsens the input graph, generating a new coarsened adjacency matrix $\mathbf{A}^{(t+1)}$ and a new matrix of coarsened embeddings $\mathbf{X}^{(t+1)}$ in this coarsened graph. In particular, *DiffGroupPool* applies the following equations:

$$\mathbf{X}^{(t+1)} = (\mathbf{M}^{(t)})^T \boldsymbol{\mu}_Z^{(t)} \in \mathbb{R}^{N\{t+1\} \times d}$$

$$\mathbf{A}^{(t+1)} = (\mathbf{M}^{(t)})^T \mathbf{A}^{(t)} \mathbf{M}^{(t)} \in \mathbb{R}^{N\{t+1\} \times N\{t+1\}}$$

## 6.4 Graph Decoder (Generator)

We denote the *tiered decoder (generator) model* as [2]:

$$p_\theta(\mathbf{A}, \mathbf{X} | \mathbf{Z}) = p_\theta(\mathbf{Z}^{(3)}) \, p_\theta(\mathbf{Z}^{(2)} | \mathbf{Z}^{(3)}) \, p_\theta(\mathbf{Z}^{(1)} | \mathbf{Z}^{(2)}) \, p_\theta(\mathbf{A}, \mathbf{X} | \mathbf{Z}^{(1)})$$

Together with the tiered encoder (inference) model, the tiered VGAE model is trained using *variational learning* [3], which maximizes the *variational lower bound*:

$$L(\mathbf{X}; \theta, \varphi) = \sum_Z q_\varphi(\mathbf{Z} | \mathbf{A}, \mathbf{X}) \log(p_\theta(\mathbf{A}, \mathbf{X} | \mathbf{Z})) - D_{KL}(q_\varphi(\mathbf{Z} | \mathbf{A}, \mathbf{X}) \| p_\theta(\mathbf{Z})),$$

and the parameters in the *tiered GNNs* are optimized by *stochastic gradient descent*.

As stated previously, currently most graph decoders are pairwise decoders, which decode pairwise relations between nodes. Developing decoding algorithms that are capable of decoding complex (sub)graphs is an important direction for future work [5]. In addition, there is no optimal method to generate graphs from embeddings [6]. We discuss some recent work in the following.

*VGAE*

The *VGAE* [11] is based on *node embeddings*. It uses an inference model:

$$q(Z \mid A, X) = \prod_{i=1}^{N} q(z_i \mid A, X), \text{ with } q(z_i \mid A, X) = \mathcal{N}(z_i \mid \mu_i, \text{diag}(\sigma_i^2)),$$

where $z_i$ is the latent variable for node i and $\mathcal{N}(.)$ is the Gaussian distribution. The inference model is parameterized by two GCNs:

$$\mu = GCN_\mu(A, X) \text{ and } \log\sigma = GCN_\sigma(A, X),$$

where $\mu$ is the matrix of $\mu_i$ and $\sigma$ is the matrix of $\sigma_i$.

The generative model is given by an inner product between latent variables:

$$p(A \mid Z) = \prod_{i=1}^{N} \prod_{j=1}^{N} p(A(i,j) \mid z_i, z_j), \text{ with } p(A(i,j) = 1 \mid z_i, z_j) = \sigma(z_i^T z_j),$$

where $\sigma(.)$ is the logistic sigmoid function. It further takes a Gaussian prior:

$$p(Z) = \prod_i p(z_i) = \prod_i \mathcal{N}(z_i \mid 0, I).$$

The VGAE is trained by maximizing the variational lower bound.

*Graphite-VAE*

The *Graphite-VAE* [15] is based on *node embeddings*. It uses an inference model:

$$q_\varphi(Z \mid A, X) = \prod_{i=1}^{N} q_\varphi(z_i \mid A, X),$$

where $z_i$ is the latent variable for node i. It further assumes isotropic Gaussian variational marginals with diagonal covariance. The inference model is parameterized by a GNN:

$$(\mu, \sigma) = GNN_\varphi(A, X).$$

The generative model is given by an inner product between latent variables:

$$p_\theta(\mathbf{A} \mid \mathbf{Z}, \mathbf{X}) = \prod_{i=1}^{N} \prod_{j=1}^{N} p_{\theta(i,j)}(\mathbf{A}(i,j) \mid \mathbf{Z}, \mathbf{X}),$$

To reconstruct the adjacency matrix $\hat{\mathbf{A}}$, an iterative two-step approach is used which alternates between defining an intermediate graph and then gradually refining this graph through message passing:

$$\hat{\mathbf{A}} = \frac{\mathbf{Z}\mathbf{Z}^T}{||\mathbf{Z}||^2} + \frac{\mathbf{1}\mathbf{1}^T}{N},$$

$$\mathbf{Z}^* = \text{GNN}_\theta(\hat{\mathbf{A}}, [\mathbf{Z}|\mathbf{X}])$$

where the second argument to the GNN is a concatenation of $\mathbf{Z}$ and $\mathbf{X}$. The two-step sequence is repeated to gradually refine $\mathbf{Z}^*$. The final $\hat{\mathbf{A}}$ is obtained using the first step on $\mathbf{Z}^*$.

The Graphite-VAE is trained by maximizing the variational lower bound.

### *CGVAE*

The *CGVAE* [16] is based on *node embeddings*. It uses an inference model:

$$q(\mathbf{Z} \mid \mathbf{A}, \mathbf{X}) = \prod_{i=1}^{N} q(\mathbf{z}_i \mid \mathbf{A}, \mathbf{X}), \quad \text{with} \quad q(\mathbf{z}_i \mid \mathbf{A}, \mathbf{X}) = \mathcal{N}(\mathbf{z}_i \mid \boldsymbol{\mu}_i, \text{diag}(\boldsymbol{\sigma}_i^2)),$$

where $\mathbf{z}_i$ is the latent variable for node i and $\mathcal{N}(.)$ is the Gaussian distribution. The inference model is parameterized by a *gated graph neural network (GGNN)* [6-7]:

$$(\boldsymbol{\mu}, \boldsymbol{\sigma}) = \text{GGNN}(\mathbf{A}, \mathbf{X}).$$

The decoder forms nodes and edges alternately. Decoding is performed by first initializing a set of possible nodes to connect. The decoder then iterates over the given nodes, performs a step of edge selection and edge labeling for the currently focused node, passes the current connected molecular graph to a GGNN for updating the node representations, and repeats this process until an edge to a special stop node is selected. This entire process is repeated for a new node in the current connected graph and terminates if there are no valid candidates.

The CGVAE is trained by maximizing the variational lower bound.

*JT-VAE*

The *JT-VAE* [17] is based on *graph embeddings* and associated *tree embeddings*. A tree decomposition is used to map a graph G into a junction tree T by contracting certain vertices into a single node so that G becomes cycle-free. In JT-VAE, both the molecular graph and its associated junction tree offer two complementary representations of molecular structure. Therefore the molecular structure is encoded into a two-part continuous latent representation $\mathbf{Z} = [\mathbf{Z}_T, \mathbf{Z}_G]$ where $\mathbf{Z}_T$ encodes the tree structure and what the subgraph components are in the tree. $\mathbf{Z}_G$ encodes the graph to capture the fine-grained connectivity.

The JT-VAE generates molecular graphs in two phases. First, it generates a tree-structured object (a junction tree) that represents the scaffold of subgraph components and their coarse relative arrangements. The components are valid chemical substructures automatically extracted from the training set using tree decomposition. Second, the subgraphs (nodes in the junction tree) are assembled into a coherent molecule graph using a graph message passing network. This approach incrementally generates the molecular graph while maintaining molecular validity at every step.

The subgraph components are used as building blocks both when encoding a molecular graph into a latent representation as well as when decoding latent representations back into valid molecular graphs. Thus, the JT-VAE encoder has two parts: graph encoder $q(\mathbf{Z}_G \mid G)$ and tree encoder $q(\mathbf{Z}_T \mid T)$, so has the decoder: tree decoder $p(T \mid \mathbf{Z}_T)$ and graph decoder $p(G \mid T, \mathbf{Z}_G)$. The graph and tree encoders are closely related to *message passing neural networks (MPNNs)* [6-7].

The tree decoder aims to maximize the likelihood $p(T \mid \mathbf{Z}_T)$. The graph decoder aims to maximize the log-likelihood of predicting correct subgraphs $SG_j$ of the ground true graph G at each tree node j.

## 7 Tiered Latent Spaces Usage and Exploration

Molecular latent representations and latent space are the center piece of deep learning for molecular structure [1], including molecular graphs. *Molecular latent space* supports interpolation, optimization and exploration, and is the *foundation* of molecular structure-property and structure-activity relationships, molecular similarity and molecular design. Tiered latent representations and latent spaces expand and increase their scope and power by providing three *disentangled but correlated tiers*, from top to bottom, the *molecule (graph) tier*, the *group tier* and the *atom (node) tier*.

## 7.1 Molecular Properties and Activities Prediction

One of the basic beliefs of chemistry is that molecular structure largely determines molecular properties or activities. *Molecular structure-property relationships* and *molecular structure-activity relationships* [1], therefore, are indispensible to molecular similarity and molecular design which depend on target molecular properties / activities. In particular, chemists can identify *groups* related to molecular properties / activities.

The tiered latent representations can be used to predict molecular properties and activities at the molecular (graph) level or at the group level, and to make atom (node) level predictions, as shown below:

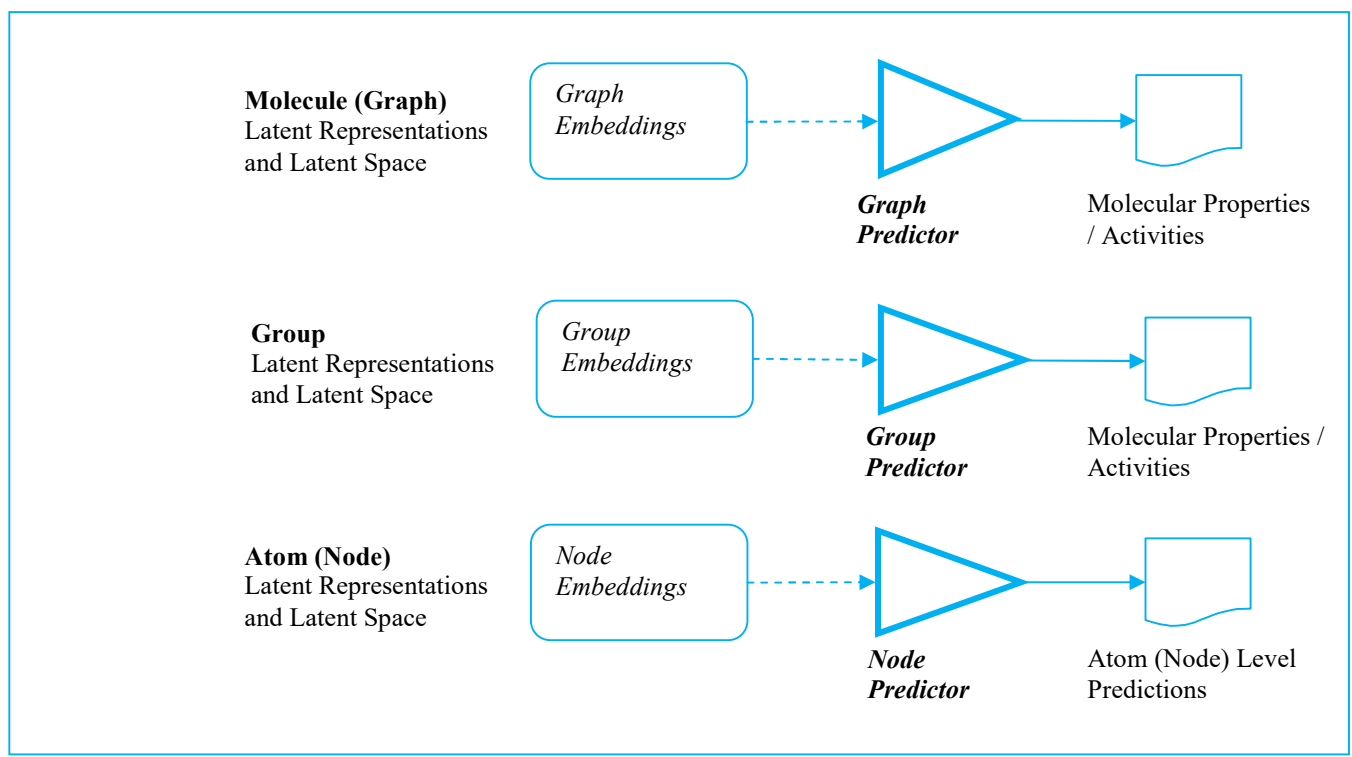

Optionally, the graph, group and/or node predictors can be jointly trained with the tiered encoders and decoder to optimize the embeddings or organizations of the latent spaces to improve prediction performance [1].

These await future investigation and exploitation.

## 7.2 Tiered Latent Spaces Exploration

The exploration of novel *chemical spaces* is one of the most important tasks of cheminformatics. The tiered latent spaces allow the exploration of novel chemical spaces at three disentangled *levels (tiers) of abstraction*, from top to bottom, the *molecule (graph) level*, the *group level* and the *atom (node) level*. Please see [2] for discussions on latent space exploration including latent space interpolation, latent space vectors and latent space geometry, as well as their significance to generative applications. Note that sampling directly from a distribution over nodes is intractable for large molecular graphs. However, sampling from graphs is tractable, and maybe even from groups.

The tiered latent spaces not only allow latent space exploration at three disentangled levels (tiers) of abstraction, but also potentially *across the levels in a coordinated manner*. This potentially can be done using the membership matrix which correlates the coarsened graphs, from top to bottom:

- $\mathbf{M}^{(2)}$: molecule (graph) <-> group
- $\mathbf{M}^{(1)}$: group <-> atom (node)

These also await future investigation and exploitation.

## 8 Summary and Conclusion

Molecular graphs generally contain groups that are identifiable and significant in composition, functionality, geometry, etc. We propose tiered latent representations and latent spaces for molecular graphs as a simple way to explicitly represent and utilize groups, which consist of the atom (node) tier, the group tier and the molecule (graph) tier. Specifically, we propose an architecture for learning tiered latent representations and latent spaces using graph autoencoders, graph neural networks, differentiable group pooling and the membership matrix.

In this paper we discussed the various components, major challenges and related work of the tiered architecture, for both a deterministic and a probabilistic model. We also briefly discussed the usage and exploration of tiered latent spaces.

For future work we plan to implement and experiment with the tiered GAE model, for molecular properties / activities prediction, as well as the tiered VGAE model, for molecular design. An important part of this will be the usage and exploration of tiered latent spaces. Other future work include: graph generation using tiered embeddings (graph, group and node) with membership matrices, and incorporation of edge features in the adjacency matrix.